\pdfoutput=1

\documentclass[11pt]{article}
\usepackage{graphicx}
\DeclareUnicodeCharacter{2717}{\texttimes}
\usepackage{acl}
\usepackage{amsmath}
\usepackage{multirow}
\usepackage{times}
\usepackage{tabularx}
\usepackage{latexsym}
\usepackage{pifont}
\usepackage{mdframed}

\usepackage{colortbl}
\usepackage[most]{tcolorbox}
\usepackage{xcolor}

\definecolor{darkred}{RGB}{255,150,150}     
\definecolor{lightred}{RGB}{255,200,200}    
\definecolor{darkyellow}{RGB}{255,230,150}  
\definecolor{lightyellow}{RGB}{255,255,200} 
\definecolor{lightgreen}{RGB}{220,255,220}  
\definecolor{darkgreen}{RGB}{180,255,180}   

\definecolor{introblue}{HTML}{4472c4}
\definecolor{introgreen}{HTML}{548235}
\definecolor{introred}{HTML}{ff0000}
\usepackage{booktabs}

\usepackage[T1]{fontenc}

\usepackage[utf8]{inputenc}

\usepackage{microtype}

\usepackage{inconsolata}

\title{\textsc{Doc2Chart}: Intent-Driven Zero-Shot Chart Generation from Documents}

\author{
  Akriti Jain, Pritika Ramu, Aparna Garimella, Apoorv Saxena \\
  Adobe Research, India \\
  \texttt{\{akritij, pramu, garimell, apoorvs\}@adobe.com}
}

\begin{document}
\maketitle
\begin{abstract} 

Large Language Models (LLMs) have demonstrated strong capabilities in transforming text descriptions or tables to data visualizations via instruction-tuning methods. 
However, it is not straightforward to apply these methods directly for a more real-world use case of visualizing data from long documents based on user-given intents, as opposed to the user pre-selecting the relevant content manually. 
We introduce the task of {\it intent-based chart generation} from documents: given a user-specified intent and document(s), the goal is to generate a chart adhering to the intent and grounded on the document(s) in a zero-shot setting. 
We propose an unsupervised, two-staged framework in which an LLM first extracts relevant information from the document(s) by decomposing the intent and iteratively validates and refines this data. 
Next, a heuristic-guided module selects an appropriate chart type before final code generation. 
To assess the data accuracy of the generated charts, we propose an attribution-based metric that uses a structured textual representation of charts, instead of relying on visual decoding metrics that often fail to capture the chart data effectively.
To validate our approach, we curate a dataset comprising of 1,242 $<$intent, document, charts$>$ tuples from two domains, finance and scientific, in contrast to the existing datasets that are largely limited to parallel text descriptions/ tables and their corresponding charts. 
We compare our approach with baselines using single-shot chart generation using LLMs and query-based retrieval methods; our method outperforms by upto $9$ points and $17$ points in terms of chart data accuracy and chart type respectively over the best baselines.
\end{abstract}

\section{Introduction}
Statistical charts offer an intuitive way to grasp insights from lengthy documents, such as financial reports and scientific articles, which are often dense with data, frequently presented in large tables. Automatic chart generation tailored to specific user goals can significantly enhance various document consumption and creation workflows. These include providing illustrative answers to specific questions, generating multi-modal summaries for queries, and creating visually rich presentations.

\begin{figure}
    \centering    \includegraphics[width=0.45\textwidth]{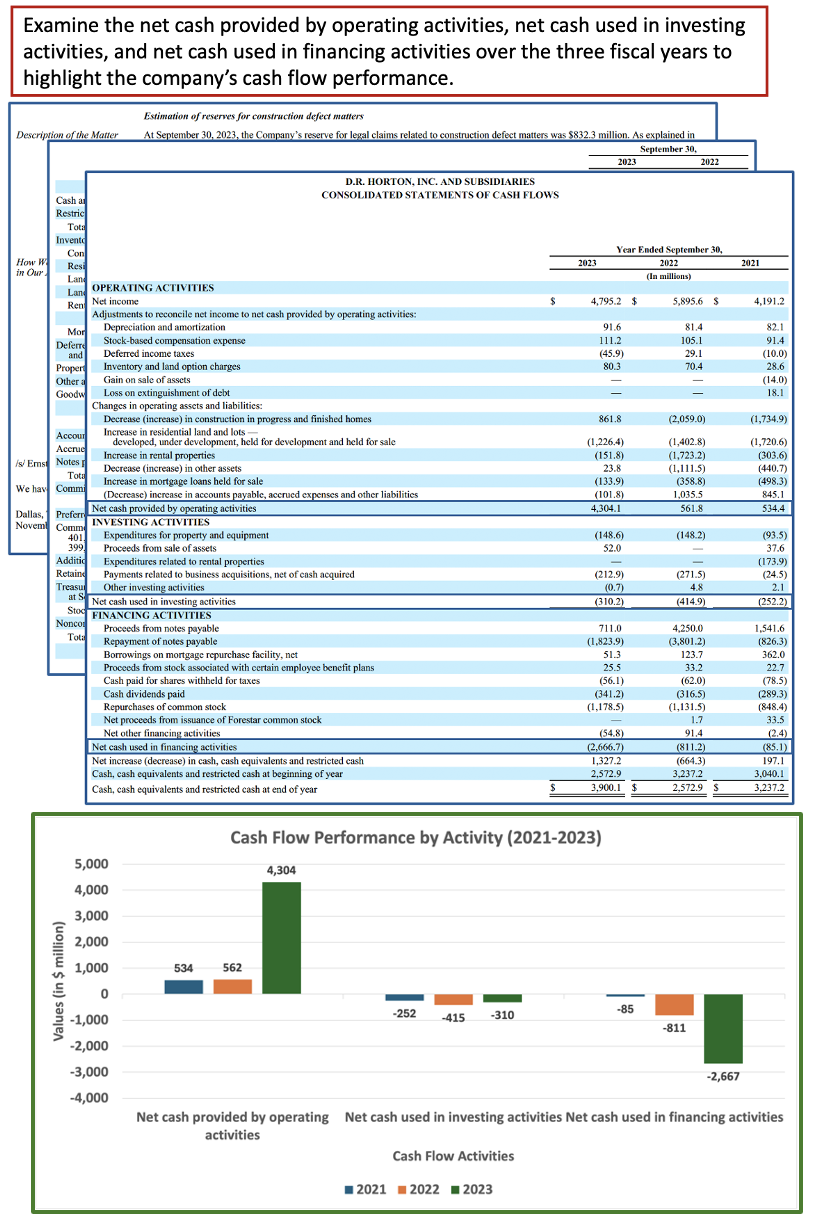}
    \caption{Example of intent-based chart generation from documents. Input: \textcolor{introred}{intent}, \textcolor{introblue}{document(s)}; output: \textcolor{introgreen}{chart}.}
    \label{fig:hero_example}
    \vspace{-0.1in}
\end{figure}
Advancements in LLMs \cite{NEURIPS2020_1457c0d6,touvron2023llama} have enabled recent efforts in generating high-quality statistical charts from user-given text descriptions or tables \cite{wang2023llm4visexplainablevisualizationrecommendation,han2023chartllamamultimodalllmchart,maddigan_chat2vis:_2023,zhang2024chartifytextautomatedchartgeneration,Tian_2024,zadeh2024text2chart31instructiontuningchart}.
However, these existing approaches to automatic chart generation typically assume that users will manually provide tables or textual descriptions to be transformed into charts. Consequently, these methods cannot directly generate charts from raw documents based on specific user intents or queries. 
Further, many previous methods rely on instruction-tuning techniques that necessitate a substantial volume of labeled data (in the order of 10s of 1000s) for effective training. 
Obtaining such large corpora of charts paired with their corresponding intents is a tedious and time-consuming process, particularly for long documents (spanning $>$ 5 pages), which are the primary focus of our work. 
 
 While LLMs demonstrate superior generation quality based on natural language prompts \cite{shao-etal-2024-assisting,ramu-etal-2024-zooming-zero}, applying them directly for chart generation using the given intent as the NL prompt often leads to charts with (a) hallucinations or poor intent adherence in terms of the data values, and (b) chart types that are not always appropriate for the given data. 
 To address these challenges, we propose an unsupervised, multi-stage framework. 
 It first employs an LLM to decompose the user’s intent to guide an iterative data extraction and refinement process from the document. Subsequently, it selects an appropriate chart type using predefined visualization heuristics before generating the final chart.
Furthermore, evaluating the fidelity of such generated charts is non-trivial. 
Existing evaluation strategies either rely on human judgments, which are costly and not easily scalable~\cite{Tian_2024, zhang2024chartifytextautomatedchartgeneration}, or employ LLM- and VLM-based evaluators~\cite{koh2024c2scalableautofeedbackllmbased, ford2024chartingfutureusingchart}, which often struggle to interpret complex charts accurately~\cite{islam2024largevisionlanguagemodels}. 
We propose an attribution-based chart evaluation metric that uses a structured text representation of generated charts, and uses attention-based heat map obtained from a forward LLM pass, to detect the spans of data values that are not captured in the reference charts.
As no existing datasets support this specific task, we finally curate a new dataset of long documents, intents, and corresponding charts from financial reports and academic papers from *ACL conferences.

This paper makes four main contributions:
{\bf (1)} We introduce the novel task of generating charts from documents based on specific intents, to mimic real-world use cases for chart generation.
{\bf (2)} We propose a training-free, multi-stage framework that leverages LLMs to first execute an intent-guided iterative data extraction and refinement process from documents, and subsequently to select an appropriate chart type through a heuristic-guided deliberation, before constructing the final chart.
{\bf (3)} We propose a chart evaluation metric to assess the faithfulness of the generated chart data when compared to either a reference chart/ table (reference-based setting) or source document itself (reference-free setting). 
{\bf (4)} We curate a dataset consisting of 1,242 $<$intent, document, chart$>$ tuples from financial reports and scientific articles for this task evaluation.

We present experimental results comparing our method with baselines including naïve prompting of LLMs and query-based retrieval methods. 
Our approach demonstrates notable improvements in key aspects such as appropriate chart type selection and chart data accuracy.
To the best of our knowledge, this is the first work to address document-to-chart generation based on intent, and can assist in furthering research in chart generation.

\section{Related Work}
Our work on intent-driven chart generation from documents intersects with several efforts, including approaches to generate and evaluate charts from more structured inputs, and techniques for understanding user intent and extracting information from documents.

\subsection{Chart Generation from Pre-Processed Inputs}
A significant body of research addresses chart generation from structured or semi-structured data.
This includes text-to-chart synthesis, where systems generate charts from concise textual descriptions \cite{rashid2021text2chartmultistagedchartgenerator, zadeh2024text2chart31instructiontuningchart} or employ reasoning over structured inputs like tables \cite{Tian_2024, wang2023llm4visexplainablevisualizationrecommendation}. Even more advanced systems that process pre-selected tabular data \cite{han2023chartllamamultimodalllmchart} or multimodal inputs \cite{xia2025chartxchartvlmversatile} primarily operate on data that is already directly present.
While these methods demonstrate strong capabilities in translating well-defined inputs into charts, they typically assume the data is already extracted, curated, and directly relevant to the desired visualization.
Our work differs significantly by tackling the upstream challenge of identifying the required information from voluminous and often complex documents based on a high-level user intent, before any chart can be synthesized.

\subsection{Intent Understanding and Information Extraction from Documents}
Understanding user intent to extract relevant information from long, complex documents is a key challenge in information retrieval and content generation. Several research directions have explored aspects of this problem. In conversational search and retrieval-augmented generation (RAG), systems such as RQ-RAG~\cite{chan2024rqraglearningrefinequeries} aim to refine ambiguous queries through user-driven clarification dialogues (e.g., “Do you mean revenue over time or per region?”). However, these approaches typically operate on short, keyword-based inputs (e.g., “revenue growth 2023”) and assume iterative user interaction. 
Unlike systems that depend on external user feedback, we incorporate internal validation and refinement steps to improve data completeness and correctness before chart generation, with the goal of producing a good-quality chart in a single pass, without requiring further clarification or user intervention. Similarly, in the table retrieval literature, recent work highlights the difficulty of locating specific table content in documents when queries are abstract or underspecified~\cite{chen2025tableretrievalsolvedproblem}. In such cases, simple keyword-matching or embedding retrieval approaches often underperform, especially when the query requires interpreting context across paragraphs and tables. While query decomposition techniques are widely used in multi-hop question answering and multi-table reasoning~\cite{chen2025tableretrievalsolvedproblem}, they are typically designed to support inference across facts rather than to extract structured components for data visualization. These methods lack fine-grained alignment to chart-specific needs, such as identifying axis labels, categories, or values. Unlike broader intent-driven generation tasks such as story writing or document drafting~\cite{shao-etal-2024-assisting,ramu-etal-2024-zooming-zero}, our problem requires the precise extraction of numerical data, grounded in the document and aligned with the user's high-level intent.

\subsection{Chart Evaluation}
Early chart evaluation works focus on comparing generated chart specifications or code to ground-truth references using measures like ROUGE, BLEU, and CodeBLEU~\cite{Tian_2024, zadeh2024text2chart31instructiontuningchart}. While these offer a measure of similarity or error, they often provide a superficial assessment, potentially overlooking semantic correctness, as data regeneration metrics alone have been found to offer a limited view of performance~\cite{ford2024chartingfutureusingchart}. Further, human evaluation, involving user studies or expert reviews \cite{Tian_2024, zhang2024chartifytextautomatedchartgeneration}, provides deeper insights into chart correctness and usability but solely relying on human evaluation can be costly and not easily scalable. More recently, automated approaches using LLMs or VLMs as evaluators ~\cite{ford2024chartingfutureusingchart} have been proposed. 
However, they often exhibit factual errors and hallucinations, particularly struggling with data extraction from charts.~\cite{islam2024largevisionlanguagemodels}.  To address these limitations while evaluating the factual accuracy in the generated charts from documents, we advocate for chart attribution: tracing the generated chart data values back to the source tables in the document.

\begin{figure*}[t]
    \centering    \includegraphics[width=1\textwidth]{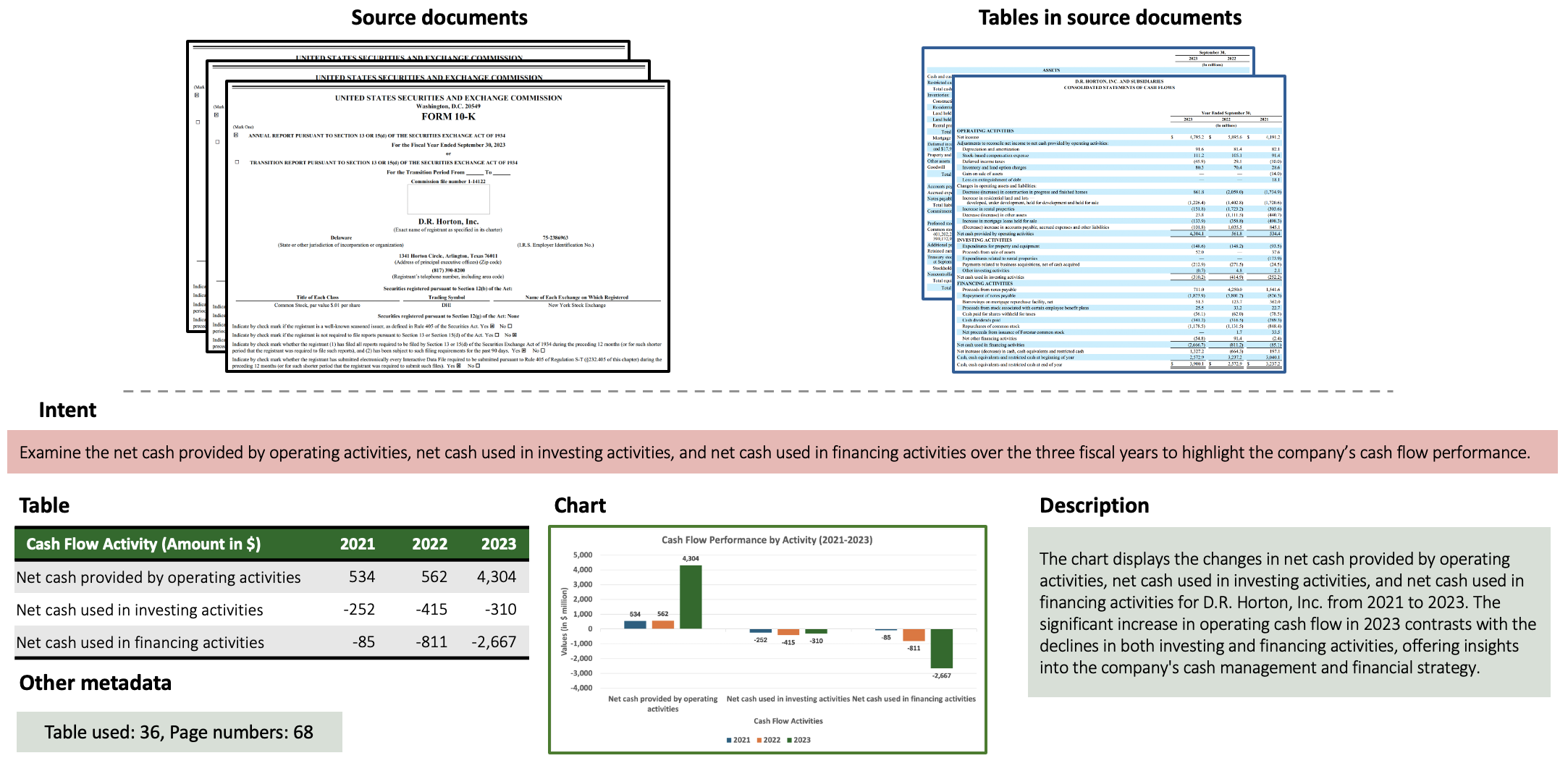}
    \caption{Sample annotations for a given SEC document and its constituent data tables. An intent and a corresponding chart are provided, along with the data table supporting the chart and a description summarizing the chart key takeaways. Other metadata such as page number and tables used from the source document are also provided. We use the page numbers to truncate the documents to include $\pm5$ pages while using the source documents with the chart generation methods, to avoid context length issues.}
    \label{fig:data_annotation_overview}
\end{figure*}
\section{Task Setup \& Dataset}
The task of intent-driven chart generation from documents is characterized by two primary challenges: first, the precise extraction of relevant data; and second, the selection of an appropriate chart type to effectively communicate this information in line with the user's intent. Consequently, for a given document \(D\) and a user-specified intent \(I\), the objective is to produce a statistical chart \(C\) that is both grounded in \(D\) and directly addresses \(I\), as depicted in Figure \ref{fig:hero_example}.
\subsection{Dataset Curation Overview}

Existing chart generation datasets primarily focus on text-to-chart \cite{rashid2021text2chartmultistagedchartgenerator,zadeh2024text2chart31instructiontuningchart} and table-to-chart \cite{han2023chartllamamultimodalllmchart} tasks, where the input text or tables contain exactly the information represented in the charts—nothing more. In contrast, our task involves generating charts from potentially long and complex documents that include significantly more content than what appears in the final visualization. To support this setting, we curate a new dataset by providing annotators with source documents (e.g., financial reports, scientific articles) and instructing them to: (a) formulate relevant user intents, and (b) create corresponding charts grounded in these documents. However, scaling such annotation is time-consuming and costly, as it requires deep ingestion of lengthy documents (often spanning tens of pages) to derive meaningful intents. This challenge motivates our adoption of a zero-shot methodology, which requires no task-specific fine-tuning.

\begin{table}[t]
\centering
\scalebox{0.8}{
\begin{tabular}{lrr}
\toprule
& \textbf{\textsc{SEC}} & \textbf{\textsc{Academic}}\\ 
\midrule
\# docs       & 73     & 106 \\ 
Avg. \# pages           & 103      & 11 \\ 
Avg. \# tables     & 24   & 6   \\ 
\# intents/ doc & 10 & 5 \\
Avg. table size  &    10 x 10&  6 x 6\\ 
\bottomrule
\end{tabular}
}
\caption{Source dataset statistics.}
\label{tab:source_data_stats}
\vspace{-0.1in}
\end{table}

For the source documents, we consider two domains, namely finance and scientific.
We use the U.S. Securities and Exchange Commission (SEC) 10-K filings\footnote{https://www.sec.gov/edgar.shtml} that are publicly available on the EDGAR website, and academic papers from *ACL conferences from the {\sc SciDuet} dataset \cite{sun-etal-2021-d2s} respectively.
We scrape the HTML documents for 1,000 SEC 10-K filings, and consider all the 1,088 papers from SciDuet in the PDF form.
We parse these HTML and PDF documents\footnote{BeautifulSoup, \url{https://developer.adobe.com/document-services/docs/overview/pdf-extract-api/}} to obtain the tables in them separately in spreadsheets. 
From among the SEC filings, we first filter documents that contain table(s) with size $>$7x7, and then pick the top 100 ones that have the maximum number of tables in them.
As most of the numeric content that can be visualized is present in tables in these documents, we use those that densely contain them.
Since academic papers do not always contain very large tables, we take the top 120 documents that have the maximum number of tables in them (and relax the size constraints).
Since these documents contain multiple tables, and each table can convey several insights, we obtain multiple intents from each of them, to reduce the number of documents that are to be ingested and the overall cognitive load.
Table \ref{tab:source_data_stats} provides the source document details for the two domains.
Figure \ref{fig:data_annotation_overview} provides an overview of the data curation process.




\begin{table}[t]
\centering
\scalebox{0.6}{
\begin{tabular}{lccrcc}
\toprule
\textbf{Dataset} & \textbf{Intent} & \textbf{Input type} & \textbf{\# Fig.} & \textbf{Desc.} &\textbf{Code}\\ 
\midrule
ChartLlama       & \ding{55}     & Table   & 11K  & \ding{51} & \ding{51}      \\ 
ChartX           & \ding{55}      & Table    & 6K  & \ding{51}  & \ding{51}  \\ 
Text2Chart31-v2     & \ding{55}   & Table  & 28.2K  & \ding{51} & \ding{51}   \\ 
\midrule
\textbf{Ours}  & \ding{51}   & \textbf{Document}  & \textbf{2.2K}   & \ding{51} & \ding{51}\\ 
\bottomrule
\end{tabular}
}
\caption{Comparative analysis with various chart generation datasets: ChartLlama \cite{han2023chartllamamultimodalllmchart}, ChartX \cite{xia2025chartxchartvlmversatile}, Text2Chart31-v2 \cite{zadeh2024text2chart31instructiontuningchart}.}
\label{tab:dataset_analysis}
\end{table}
\begin{figure*}[t]
    \centering    \includegraphics[width=\textwidth]{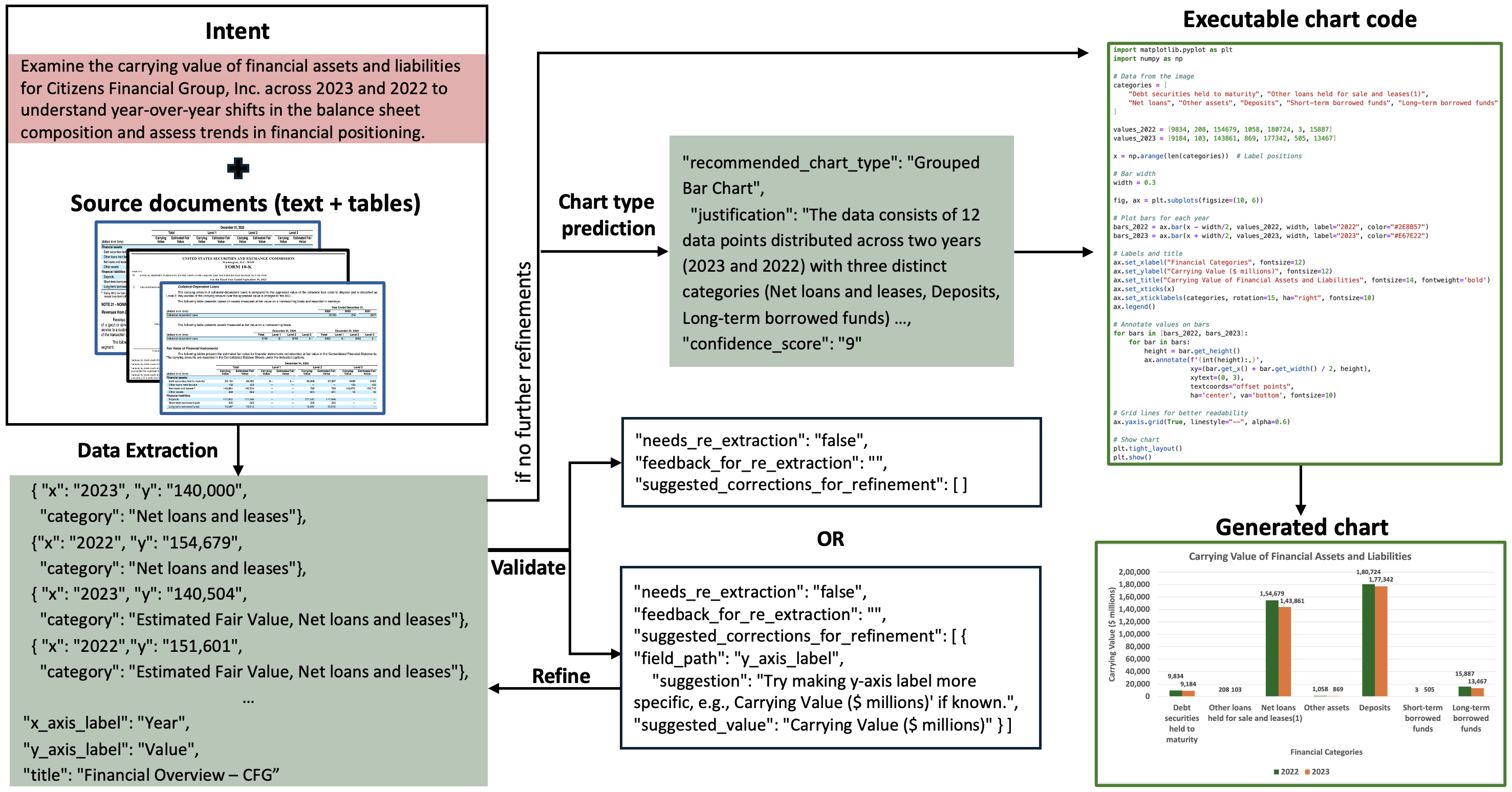}
    \caption{The proposed pipeline, illustrating: tterative data extraction, where misalignment with intent or data incompleteness facilitates a re-extraction. A refinement stage, where minor issues in otherwise suitable data are corrected. Chart type prediction based on the finalized data, followed by code generation }
    \label{fig:method_overview}
\end{figure*}
\subsection{Data Annotation Setup}
{To obtain intent-chart annotations, we recruited three annotators from a freelancing platform,\footnote{\url{https://www.upwork.com}} all proficient in content creation with similar demographics (nationality, graduate education, age 20-30). After pilot studies (5 documents each), they were instructed to provide: (a) 5 creative intents; (b) corresponding chart images and tables with appropriate data and chart types; (c) supporting source content (e.g., tables, page numbers); and (d) text descriptions of the charts. To account for subjectivity, annotators could list up to three chart type variants per intent, ordered by preference. For large tables, they were encouraged to combine data from multiple tables or use subsets, critically considering insightful visualizations. Following 2-3 feedback rounds focused on ensuring intents were neither too generic nor specific, the main task involved generating 5 intent-chart pairs for academic articles and 10 for SEC filings, initially yielding 1,275 data points. The authors, as expert reviewers, then filtered samples with ambiguous intents or inaccurate charts, resulting in 1,242 <intent, document, chart> tuples. Table \ref{tab:dataset_analysis} compares our dataset with three existing chart generation datasets. Including plausible chart type variants for some samples increased the total chart count to 2.2K charts.}
The annotators are compensated with \$15/ hour.

\section{\textsc{Doc2Chart}: Methodology}

Generating charts directly from documents using user's intent as a simple prompt for a Large Language Model (LLM) often yields suboptimal results. Such charts can suffer from poor intent adherence, include hallucinated or irrelevant data, or omit crucial information, especially when data spans multiple segments or requires subsetting from large tables in long documents. 
To address these issues, we propose \textsc{Doc2Chart}, an unsupervised, multi-stage framework (illustrated in Figure \ref{fig:method_overview}). Our pipeline systematically processes the document and intent to produce an accurate and appropriate chart. 

\paragraph{Iterative Data Extraction and Refinement.}
The first stage involves decomposing the user intent and identifying relevant content from the document. To ensure the highest fidelity of the extracted data when dealing with the complexities of long-form content, this initial extraction undergoes a critical validation and refinement phase. This step systematically verifies the accuracy, completeness, and relevance of the information, applying corrections or guiding re-extraction as needed. The outcome of this validation dictates the next step. If critical issues such as significant data omissions or fundamental misinterpretations of the intent are identified, the validation module generates specific feedback. This feedback then guides a subsequent extraction attempt, repeating the initial extraction step. If no corrections are deemed necessary, the validated data moves forward directly. This iterative cycle of extraction, validation, and conditional re-extraction or refinement ensures a high degree of alignment between the extracted data, the source document, and the user's intent.

\paragraph{Chart Type Prediction.}
After verifying and refining the data, the next step is to determine the most suitable chart type. This choice can be non-trivial, as multiple chart types may fit a given dataset and intent. We adaopt a heuristic-guided approach where an LLM analyzes the structure of the data—such as the types of values and the number of data points or categories, in conjunction with the user's intent and recommends a chart type accordingly. 
For example, line charts are typically preferred for time-series data to highlight trends, while bar charts may be used when the data contains only a few points. Simple categorical comparisons suit standard bar charts, whereas grouped or stacked bar charts are better for subcategory comparisons. Pie charts can be effective for part-to-whole relationships, but only when the number of segments is small enough to remain clear. 
These also help the model avoid common visualization pitfalls, such as cluttered visuals or misleading representations. 
Rather than relying on rigid rules, the LLM combines data characteristics with these heuristics to recommend a chart type, along with a justification and a confidence score for its choice.

\paragraph{Code Generation.}
Finally, we generate an executable chart code with \texttt{Matplotlib} based on the extracted data and the selected chart type for rendering the chart.

\begin{table*}[t]
\centering
\scalebox{0.75}{
\setlength{\tabcolsep}{6pt}
\resizebox{\textwidth}{!}{
\begin{tabular}{llccc}
\toprule
\textbf{Model} & \textbf{Method} & {\textbf{Chart Data}} & \multicolumn{2}{c}{\textbf{Chart Type}} \\
\textbf{} & \textbf{} & \textbf{} & \textbf{Best} & \textbf{Out-of-3} \\
\midrule

\multirow{6}{*}{GPT-4o} 
    & Single-step & \cellcolor{lightred}67.38& \cellcolor{lightred}71.79& \cellcolor{lightred}75.63\\
\cmidrule(lr){2-5}
    & Embedding retrieval & 38.98& 35.04& 38.97\\
    & LLM retrieval & 59.97& 62.90& 68.11\\
    & LLM retrieval (w/ ques decomp) & 57.09& 68.45& 73.44\\
\cmidrule(lr){2-5}
    & \textbf{\textsc{Doc2Chart}} & \cellcolor{lightgreen}75.18& \cellcolor{lightgreen}79.49& \cellcolor{lightgreen}82.62\\
\midrule

\multirow{6}{*}{Gemini-2.0}  
    & Single-step & \cellcolor{lightred}62.51& 49.36& 52.28\\
\cmidrule(lr){2-5}
    & Embedding retrieval & 38.92& 14.71 & 14.86 \\
    & LLM retrieval & 59.21& 45.55& 48.63\\
    & LLM retrieval (w/ ques decomp) & 50.01& \cellcolor{lightred}59.72& \cellcolor{lightred}63.48\\
\cmidrule(lr){2-5}
    & \textbf{\textsc{Doc2Chart}} & \cellcolor{lightgreen}71.53& \cellcolor{lightgreen}74.12& \cellcolor{lightgreen}79.41\\
\midrule

\multirow{6}{*}{Claude-3.5-Sonnet}  
    & Single-step & 63.75& \cellcolor{lightred}64.80& \cellcolor{lightred}67.92\\
\cmidrule(lr){2-5}
    & Embedding retrieval & 43.69 & 27.29& 28.28\\
    & LLM retrieval & 58.60& 61.50& 66.33\\
    & LLM retrieval (w/ ques decomp) & \cellcolor{lightred}64.48 & 61.27& 64.74\\
\cmidrule(lr){2-5}
    & \textbf{\textsc{Doc2Chart}} & \cellcolor{lightgreen}69.45& \cellcolor{lightgreen}82.01& \cellcolor{lightgreen}84.13\\
\midrule

\multirow{6}{*}{LLaMA-3.1-8B-Instruct} 
    & Single-step & 39.20& \cellcolor{lightred}54.47& \cellcolor{lightred}58.30\\
\cmidrule(lr){2-5}
    & Embedding retrieval & 24.09& 15.66 & 17.78 \\
    & LLM retrieval & 27.11& 40.67 & 43.31 \\
    & LLM retrieval (w/ ques decomp) & \cellcolor{lightred}39.74& 48.98& 52.48\\
\cmidrule(lr){2-5}
    & \textbf{\textsc{Doc2Chart}} & \cellcolor{lightgreen}42.17& \cellcolor{lightgreen}71.75& \cellcolor{lightgreen}78.05\\
\bottomrule

\end{tabular}}}
\caption{Performance comparison of various methods on chart data accuracy and chart type selection. \textcolor{lightgreen}{Green} highlights best performance; \textcolor{lightred}{Red} highlights second-best. Our methods \textsc{Doc2Chart} consistently outperform baselines.}
\label{tab:auto_evaluation}
\vspace{-0.1in}
\end{table*}

\section{Experimental Setup}
We conduct experiments using four LLMs, namely GPT-4o \cite{openai2024gpt4technicalreport}, Gemini-2.0 \cite{gemini2024flash}, LLaMA-3.1-8B-Instruct \cite{meta2024llama3} and Claude-3.5-Sonnet \cite{noauthor_introducing_nodate}.
\subsection{Baselines} We compare our approach against four baselines. \\  
\noindent
\textbf{Single-Step Generation} serves as the most straightforward approach, directly generating the chart from the input document and user intent without any intermediate retrieval.  \\
\noindent
\textbf{Embedding-Based Retrieval} incorporates a retrieval step as no parallel data tables are available for the intents. The document is segmented based on headings, and SBERT embeddings (all-MiniLM-L6-v2) \cite{reimers-2019-sentence-bert} are used to retrieve sections most relevant to the intent which are then used for chart code generation. \\ 
\noindent
\textbf{LLM-Based Retrieval} builds on recent advances in LLM-powered retrieval \cite{zhu2024largelanguagemodelsinformation}, which have demonstrated that LLMs can outperform embedding-based approaches by capturing richer contextual relationships between queries and documents. In this baseline, an LLM is used as a retriever before generating the chart code.  \\
\noindent
\textbf{Query Decomposition for Table Retrieval} takes advantage of the fact that most chart content comes from tables. Inspired by \citet{chen2025tableretrievalsolvedproblem}, this method decomposes intents into (concept, attribute) pairs to enhance retrieval accuracy. Query decomposition is first applied, followed by LLM-based retrieval to extract relevant tabular data before generating the chart code.  

In all the baselines, the chart type prediction is fixed to a naïve LLM instruction for a suitable type.

\subsection{Evaluation Metrics}
The generated charts should accurately reflect the underlying data from the documents and use appropriate chart types to convey the intended information.
To evaluate these aspects, we use: (a) {\it chart data accuracy}, which includes completeness, correctness, and overall data quality with respect to the reference chart and (b) {\it chart type validation}, to compare the predicted type with the ground-truth chart types. 
For data accuracy, recent works either use n-gram text similarity metrics for the generated code \cite{han2023chartllamamultimodalllmchart,zadeh2024text2chart31instructiontuningchart} or Visual Language Models (VLMs) to compare the generated and reference charts.
However, such n-gram-based measures are known to be limited to surface-level aspects and fail to capture the nuanced error cases in the data values. 
VLMs often struggle with faithfully interpreting complex charts, especially when dealing with multiple data points or subtle variations in values \cite{huang-etal-2024-lvlms}, and are inherently limited by their visual perception accuracy \cite{ford2024chartingfutureusingchart}.

We take inspiration in attribution as a strategy to validate specific spans of text by grounding them in the source context, both in textual space and more recently for VLMs \cite{jiang2025interpretingeditingvisionlanguagerepresentations, phukan2025logitlenscontextualembeddings} 
We propose \textbf{\textsc{ChartEval}}, an attribution-based metric that uses structured textual representations for charts, and avoids reliance on visual decoding altogether. 
We trace the intermediate representations of the generated charts, extracted as structured JSONs before code generation, back to their source tables in the document(s). 
Each chart is represented as a collection of tuples $\langle x\text{-axis}, y\text{-axis}, \text{value} \rangle$ in the JSON, each of which is individually validated against the values in the corresponding tables in ground truth. 
For this validation, we use a modified version of the attribution algorithm from \cite{phukan2024peeringmindlanguagemodels,cohenwang2025learningattributeattention}: {\it (i)} We construct a prompt using the reference table\footnote{This can be used for reference chart as well by obtaining its JSON representation, if table references are not available.} as the "document", and generated chart JSON as the "output".
{\it (ii)} We forward pass this prompt through the Llama-3.1-8b-Instruct model and aggregate the cross attention scores between the "output" and "document" tokens to get a token-level heatmap of $<$output tokens, document tokens$>$ size.
{\it (iii)} For each data value token in the "output", we identify the best matching span in the "document" tokens using Kadane's algorithm \cite{kadane2023} on the obtained heatmap.
While this formulation of {\sc ChartEval} does a reference-based evaluation with the ground truth tables, it can be extended to a reference-free variant as well, where the "document" would be the entire source document context in the LLM forward pass.

\subsection{Human Evaluation for Metric Quality}

To qualitatively validate our approach, we conduct human surveys to compare the model-generated charts against references. 
Each evaluation instance consists of an intent, ground truth (GT) chart, its corresponding GT table, along with three AI-generated charts (from our approach and two baselines), which are anonymized and randomized to eliminate any biases. 
We hire three expert annotators and provide them with 300 instances (150 from each domain) to rate them on six criteria:
\textit{Chart Data Correctness} to measure whether all values in the AI-generated chart match the reference data exactly;
\textit{Chart Data Completeness} to evaluate whether the chart includes all relevant values from the reference table;
\textit{Overall Chart Data Quality} to determine how accurately the chart conveys the reference data; 
\textit{Chart Type Validation} to check if the selected chart type aligns with the given intent and data;
\textit{Chart Insightfulness} to assess how well the chart highlights key insights, adheres to intent, and whether visual encoding (e.g., colors, labels, legends) aids understanding;
finally, \textit{Overall Chart Quality} to assess the clarity, accuracy, and usefulness of the generated chart. Each criterion is rated on a four-point scale: \textit{Minimal}, \textit{Partial}, \textit{Most}, and \textit{Full}, where \textit{Minimal} indicates the chart does not meet expectations at all, and \textit{Full} represents an ideal chart. 
If a chart only partially contains correct values—either missing some or including incorrect ones—the rating is to be adjusted accordingly.
We compute correlations between {\sc ChartEval} ratings and human ratings on these 900 samples, and find a strong alignment with human judgments (Pearson’s $r=0.71$).
Using a simpler LLM-based metric, on the other hand, where the generated chart tuples and reference table values are given to an LLM which is then instructed to provide a rating for data accuracy, we note a much lower $r=0.39$.


\section{Results \& Discussion}

\begin{table}[t]
    \centering
    \scriptsize
    \renewcommand{\arraystretch}{1.1}
    \setlength{\tabcolsep}{5pt}
    \begin{tabular}{lccc}
        \toprule
        \textbf{Metric} & \textbf{Single-Step} & \textbf{LLM-R w/ QD } & \textbf{Ours} \\
        \midrule
        Chart Data Correctness      &  2.98&  2.59&  \textbf{3.52}\\
        Chart Data Completeness     &  3.12&  2.66&  \textbf{3.63}\\
        Overall Data Quality        &  2.98&  2.59&  \textbf{3.53}\\
        Chart Type Validation       &  3.72&  3.33&  \textbf{4.13}\\
        Insightfulness              &  2.77&  2.43&  \textbf{3.36}\\
        \midrule
        \textbf{Overall Quality}    &  2.74&  2.41&  \textbf{3.34}\\
        \bottomrule
    \end{tabular}
    \caption{Comparison of different methods against human ratings across evaluation axes. LLM-R w/ QD: LLM retrieval with query decomposition.}
    \vspace{-0.1in}
    \label{tab:human_evaluation}
\end{table}

We evaluate the performance of {\sc Doc2Chart} pipeline using four LLMs: GPT-4o, Gemini-2.0, Claude-3.5-Sonnet, and LLaMA-3.1-8B-Instruct (Table \ref{tab:auto_evaluation}). 
\noindent In terms of chart data accuracy, \textsc{Doc2Chart} significantly outperforms all baselines. 
For instance, with GPT-4o, \textsc{Doc2Chart} achieves $75.18\%$ accuracy, a notable improvement over that for the single-step baseline ($67.38\%$) and other retrieval-augmented approaches such as LLM retrieval (w/ ques decomp) ($57.09\%$). 
Similar trends are observed for other models; with Gemini-2.0, \textsc{Doc2Chart} ($71.53\%$) substantially surpasses LLM retrieval (w/ ques decomp) ($50.01\%$).
This underscores the benefit of the iterative refinement process in improving the factual correctness of the extracted data. 
Similar gains are observed in chart-type prediction across models using the heuristic-based analysis using LLM, compared to naïve prompting. 
The results consistently show that improving chart data accuracy through validation and refinement directly translates to better chart type recommendations as well. 


Among the baselines, the single-step one performs the best in most cases for data accuracy. Embedding-based retrieval consistently underperforms, lacking the necessary contextual depth for accurate chart data extraction. While LLM-based retrieval, especially when combined with query decomposition, shows some improvement over simple embeddings, it still lags behind. 
The query decomposition baseline struggles as it only breaks down the intent into broad topics and attributes rather than structured data tuples. 
For example, when given the intent "{\it Assess the hotel revenues for the top 5 highest
performing regions from 2021 to 2023, focusing on the trends in revenue growth and regional performance,}" it outputs generic components like \texttt{<sub\_c>hotels:revenue</sub\_c>} and \texttt{<sub\_c>revenue:trend</sub\_c>}. However, these lack the necessary structure to retrieve precise data. 
Table \ref{tab:human_evaluation} shows the human ratings for the generations using our approach and two other baselines (taking majority rating for each sample and averaging across samples). Our method's outputs are consistently rated higher than those generated by the baselines, and the single-step baselines are rated as the next best.

\section{Conclusions \& Future Work}
We present the task of intent-based chart generation from documents, where the objective is to generate charts that not only align with a user-specified intent but are also grounded in the source document. In contrast to prior datasets that focus on table-to-chart or plain text-to-chart generation, our dataset includes 1,242 (intent, document, chart) tuples, reflecting more realistic and open-ended scenarios. Our unsupervised, multi-stage framework decomposes the user intent to iteratively extract and refine relevant data,selected appropriate chart type using visualization heuristics, and generates executable chart code in a zero-shot manner. While ideal data accuracy would be close to 1—especially to assist users in bypassing the need to manually navigate long documents—our method consistently outperforms strong baselines, including single-shot generation and retrieval-based methods, across both data accuracy and chart type selection. To increase trust in AI-generated charts, we advocate for chart attribution: tracing chart values back to their textual sources. Future work can extend this by attributing data values not just to tables but to specific text spans. Attribution failures may also serve as useful feedback signals to guide data refinement. While human evaluations reflect similar trends, it is important to note that chart type selection can be inherently subjective—multiple chart types may be valid depending on the user’s analytic goal—so future evaluation strategies should account for this flexibility.
We hope that our work paves the way towards developing more accessible, intent-driven document visualizations, with potential applications in domains like finance, science, and public policy.

\section{Limitations}
User intents can often be vague or underspecified, leading to multiple valid interpretations. While our framework performs intent decomposition and iterative refinement to approximate the user's needs, it does not incorporate dynamic user feedback or interactive clarification. Although interactive mechanisms—where users confirm or adjust extracted information—could further improve alignment with user expectations, our focus is on generating a high-quality first draft without requiring user intervention. Additionally, due to the limited context window of current LLMs, scaling the approach to handle multiple long documents remains a challenge. 
Lastly, while our chart attribution metric helps evaluate factual grounding, it is currently implemented in a reference-based setting—comparing chart data directly against source tables. This metric can be extended to a reference-free setup, where the attribution model takes the generated table and raw document markdown(s) as input. While this improves scalability, it may occasionally attribute values to the wrong context if the same value appears elsewhere in the document—a challenge we leave for future work.

\section{Ethical Statement}
Automated chart generation carries the risk of producing misleading visualizations, especially in high-stakes domains such as finance, science, and policy. To mitigate these risks, our work emphasizes faithfulness and transparency by introducing chart attribution—a method that traces chart content back to its source tables—alongside quantitative evaluation grounded in document data. We also avoid generating speculative content and restrict chart construction to source-supported values only. Nonetheless, we acknowledge that LLMs may still produce flawed outputs (hallucinations). Future work should explore mechanisms for uncertainty estimation, user-in-the-loop validation, and better safeguards to ensure responsible deployment of automatic charting systems.

\bibliography{custom}

\begin{thebibliography}{30}
\expandafter\ifx\csname natexlab\endcsname\relax\def\natexlab#1{#1}\fi

\bibitem[{Anthropic()}]{noauthor_introducing_nodate}
Anthropic.
\newblock \href {https://www.anthropic.com/news/claude-3-5-sonnet} {Introducing {Claude} 3.5 {Sonnet}}.

\bibitem[{Brown et~al.(2020)Brown, Mann, Ryder, Subbiah, Kaplan, Dhariwal, Neelakantan, Shyam, Sastry, Askell, Agarwal, Herbert-Voss, Krueger, Henighan, Child, Ramesh, Ziegler, Wu, Winter, Hesse, Chen, Sigler, Litwin, Gray, Chess, Clark, Berner, McCandlish, Radford, Sutskever, and Amodei}]{NEURIPS2020_1457c0d6}
Tom Brown, Benjamin Mann, Nick Ryder, Melanie Subbiah, Jared~D Kaplan, Prafulla Dhariwal, Arvind Neelakantan, Pranav Shyam, Girish Sastry, Amanda Askell, Sandhini Agarwal, Ariel Herbert-Voss, Gretchen Krueger, Tom Henighan, Rewon Child, Aditya Ramesh, Daniel Ziegler, Jeffrey Wu, Clemens Winter, Chris Hesse, Mark Chen, Eric Sigler, Mateusz Litwin, Scott Gray, Benjamin Chess, Jack Clark, Christopher Berner, Sam McCandlish, Alec Radford, Ilya Sutskever, and Dario Amodei. 2020.
\newblock \href {https://proceedings.neurips.cc/paper_files/paper/2020/file/1457c0d6bfcb4967418bfb8ac142f64a-Paper.pdf} {Language models are few-shot learners}.
\newblock In \emph{Advances in Neural Information Processing Systems}, volume~33, pages 1877--1901. Curran Associates, Inc.

\bibitem[{Chan et~al.(2024)Chan, Xu, Yuan, Luo, Xue, Guo, and Fu}]{chan2024rqraglearningrefinequeries}
Chi-Min Chan, Chunpu Xu, Ruibin Yuan, Hongyin Luo, Wei Xue, Yike Guo, and Jie Fu. 2024.
\newblock \href {http://arxiv.org/abs/2404.00610} {Rq-rag: Learning to refine queries for retrieval augmented generation}.

\bibitem[{Chen et~al.(2025)Chen, Zhang, and Roth}]{chen2025tableretrievalsolvedproblem}
Peter~Baile Chen, Yi~Zhang, and Dan Roth. 2025.
\newblock \href {http://arxiv.org/abs/2404.09889} {Is table retrieval a solved problem? exploring join-aware multi-table retrieval}.

\bibitem[{Cohen-Wang et~al.(2025)Cohen-Wang, Chuang, and Madry}]{cohenwang2025learningattributeattention}
Benjamin Cohen-Wang, Yung-Sung Chuang, and Aleksander Madry. 2025.
\newblock \href {http://arxiv.org/abs/2504.13752} {Learning to attribute with attention}.

\bibitem[{Ford et~al.(2024)Ford, Zhao, Schumacher, and Rios}]{ford2024chartingfutureusingchart}
James Ford, Xingmeng Zhao, Dan Schumacher, and Anthony Rios. 2024.
\newblock \href {http://arxiv.org/abs/2409.18764} {Charting the future: Using chart question-answering for scalable evaluation of llm-driven data visualizations}.

\bibitem[{Google(2024)}]{gemini2024flash}
Google. 2024.
\newblock \href {https://cloud.google.com/vertex-ai/generative-ai/docs/gemini-v2} {Gemini 2.0 flash}.

\bibitem[{Han et~al.(2023)Han, Zhang, Chen, Yang, Wang, Yu, Fu, and Zhang}]{han2023chartllamamultimodalllmchart}
Yucheng Han, Chi Zhang, Xin Chen, Xu~Yang, Zhibin Wang, Gang Yu, Bin Fu, and Hanwang Zhang. 2023.
\newblock \href {http://arxiv.org/abs/2311.16483} {Chartllama: A multimodal llm for chart understanding and generation}.

\bibitem[{Huang et~al.(2024)Huang, Zhou, Chan, Fung, Wang, Zhang, Chang, and Ji}]{huang-etal-2024-lvlms}
Kung-Hsiang Huang, Mingyang Zhou, Hou~Pong Chan, Yi~Fung, Zhenhailong Wang, Lingyu Zhang, Shih-Fu Chang, and Heng Ji. 2024.
\newblock \href {https://doi.org/10.18653/v1/2024.findings-acl.41} {Do {LVLM}s understand charts? analyzing and correcting factual errors in chart captioning}.
\newblock In \emph{Findings of the Association for Computational Linguistics: ACL 2024}, pages 730--749, Bangkok, Thailand. Association for Computational Linguistics.

\bibitem[{Islam et~al.(2024)Islam, Rahman, Masry, Laskar, Nayeem, and Hoque}]{islam2024largevisionlanguagemodels}
Mohammed~Saidul Islam, Raian Rahman, Ahmed Masry, Md~Tahmid~Rahman Laskar, Mir~Tafseer Nayeem, and Enamul Hoque. 2024.
\newblock \href {http://arxiv.org/abs/2406.00257} {Are large vision language models up to the challenge of chart comprehension and reasoning? an extensive investigation into the capabilities and limitations of lvlms}.

\bibitem[{Jiang et~al.(2025)Jiang, Kachinthaya, Petryk, and Gandelsman}]{jiang2025interpretingeditingvisionlanguagerepresentations}
Nick Jiang, Anish Kachinthaya, Suzie Petryk, and Yossi Gandelsman. 2025.
\newblock \href {http://arxiv.org/abs/2410.02762} {Interpreting and editing vision-language representations to mitigate hallucinations}.

\bibitem[{Kadane(2023)}]{kadane2023}
Joseph~B. Kadane. 2023.
\newblock \href {https://doi.org/10.3390/algorithms16110519} {Two kadane algorithms for the maximum sum subarray problem}.
\newblock \emph{Algorithms}, 16(11):519.

\bibitem[{Koh et~al.(2024)Koh, Yoon, Lee, Song, Cho, Kang, Kim, young Yun, Yu, and Lee}]{koh2024c2scalableautofeedbackllmbased}
Woosung Koh, Jang~Han Yoon, MinHyung Lee, Youngjin Song, Jaegwan Cho, Jaehyun Kang, Taehyeon Kim, Se~young Yun, Youngjae Yu, and Bongshin Lee. 2024.
\newblock \href {http://arxiv.org/abs/2410.18652} {$c^2$: Scalable auto-feedback for llm-based chart generation}.

\bibitem[{Maddigan and Susnjak(2023)}]{maddigan_chat2vis:_2023}
Paula Maddigan and Teo Susnjak. 2023.
\newblock \href {https://doi.org/10.1109/ACCESS.2023.3274199} {{Chat2VIS}: {Generating} {Data} {Visualizations} via {Natural} {Language} {Using} {ChatGPT}, {Codex} and {GPT}-3 {Large} {Language} {Models}}.
\newblock \emph{IEEE Access}, 11:45181--45193.

\bibitem[{Meta(2024)}]{meta2024llama3}
Meta. 2024.
\newblock \href {https://arxiv.org/abs/2407.21783} {The llama 3 herd of models}.
\newblock \emph{arXiv preprint arXiv:2407.21783}.

\bibitem[{OpenAI(2024)}]{openai2024gpt4technicalreport}
OpenAI. 2024.
\newblock \href {http://arxiv.org/abs/2303.08774} {Gpt-4 technical report}.

\bibitem[{Phukan et~al.(2025)Phukan, Divyansh, Morj, Vaishnavi, Saxena, and Goswami}]{phukan2025logitlenscontextualembeddings}
Anirudh Phukan, Divyansh, Harshit~Kumar Morj, Vaishnavi, Apoorv Saxena, and Koustava Goswami. 2025.
\newblock \href {http://arxiv.org/abs/2411.19187} {Beyond logit lens: Contextual embeddings for robust hallucination detection \& grounding in vlms}.

\bibitem[{Phukan et~al.(2024)Phukan, Somasundaram, Saxena, Goswami, and Srinivasan}]{phukan2024peeringmindlanguagemodels}
Anirudh Phukan, Shwetha Somasundaram, Apoorv Saxena, Koustava Goswami, and Balaji~Vasan Srinivasan. 2024.
\newblock \href {http://arxiv.org/abs/2405.17980} {Peering into the mind of language models: An approach for attribution in contextual question answering}.

\bibitem[{Ramu et~al.(2024)Ramu, Gaur, Emandi, Maheshwari, Javed, and Garimella}]{ramu-etal-2024-zooming-zero}
Pritika Ramu, Pranshu Gaur, Rishita Emandi, Himanshu Maheshwari, Danish Javed, and Aparna Garimella. 2024.
\newblock \href {https://aclanthology.org/2024.inlg-main.52/} {Zooming in on zero-shot intent-guided and grounded document generation using {LLM}s}.
\newblock In \emph{Proceedings of the 17th International Natural Language Generation Conference}, pages 676--694, Tokyo, Japan. Association for Computational Linguistics.

\bibitem[{Rashid et~al.(2021)Rashid, Jahan, Huzzat, Rahul, Zakir, Meem, Mukta, and Shatabda}]{rashid2021text2chartmultistagedchartgenerator}
Md.~Mahinur Rashid, Hasin~Kawsar Jahan, Annysha Huzzat, Riyasaat~Ahmed Rahul, Tamim~Bin Zakir, Farhana Meem, Md. Saddam~Hossain Mukta, and Swakkhar Shatabda. 2021.
\newblock \href {http://arxiv.org/abs/2104.04584} {Text2chart: A multi-staged chart generator from natural language text}.

\bibitem[{Reimers and Gurevych(2019)}]{reimers-2019-sentence-bert}
Nils Reimers and Iryna Gurevych. 2019.
\newblock \href {https://arxiv.org/abs/1908.10084} {Sentence-bert: Sentence embeddings using siamese bert-networks}.
\newblock In \emph{Proceedings of the 2019 Conference on Empirical Methods in Natural Language Processing}. Association for Computational Linguistics.

\bibitem[{Shao et~al.(2024)Shao, Jiang, Kanell, Xu, Khattab, and Lam}]{shao-etal-2024-assisting}
Yijia Shao, Yucheng Jiang, Theodore Kanell, Peter Xu, Omar Khattab, and Monica Lam. 2024.
\newblock \href {https://doi.org/10.18653/v1/2024.naacl-long.347} {Assisting in writing {W}ikipedia-like articles from scratch with large language models}.
\newblock In \emph{Proceedings of the 2024 Conference of the North American Chapter of the Association for Computational Linguistics: Human Language Technologies (Volume 1: Long Papers)}, pages 6252--6278, Mexico City, Mexico. Association for Computational Linguistics.

\bibitem[{Sun et~al.(2021)Sun, Hou, Wang, Zhang, and Wang}]{sun-etal-2021-d2s}
Edward Sun, Yufang Hou, Dakuo Wang, Yunfeng Zhang, and Nancy X.~R. Wang. 2021.
\newblock \href {https://doi.org/10.18653/v1/2021.naacl-main.111} {{D}2{S}: Document-to-slide generation via query-based text summarization}.
\newblock In \emph{Proceedings of the 2021 Conference of the North American Chapter of the Association for Computational Linguistics: Human Language Technologies}, pages 1405--1418, Online. Association for Computational Linguistics.

\bibitem[{Tian et~al.(2024)Tian, Cui, Deng, Yi, Yang, Zhang, and Wu}]{Tian_2024}
Yuan Tian, Weiwei Cui, Dazhen Deng, Xinjing Yi, Yurun Yang, Haidong Zhang, and Yingcai Wu. 2024.
\newblock \href {https://doi.org/10.1109/tvcg.2024.3368621} {Chartgpt: Leveraging llms to generate charts from abstract natural language}.
\newblock \emph{IEEE Transactions on Visualization and Computer Graphics}, page 1–15.

\bibitem[{Touvron et~al.(2023)Touvron, Martin, Stone, Albert, Almahairi, Babaei, Bashlykov, Batra, Bhargava, Bhosale, Bikel, Blecher, Ferrer, Chen, Cucurull, Esiobu, Fernandes, Fu, Fu, Fuller, Gao, Goswami, Goyal, Hartshorn, Hosseini, Hou, Inan, Kardas, Kerkez, Khabsa, Kloumann, Korenev, Koura, Lachaux, Lavril, Lee, Liskovich, Lu, Mao, Martinet, Mihaylov, Mishra, Molybog, Nie, Poulton, Reizenstein, Rungta, Saladi, Schelten, Silva, Smith, Subramanian, Tan, Tang, Taylor, Williams, Kuan, Xu, Yan, Zarov, Zhang, Fan, Kambadur, Narang, Rodriguez, Stojnic, Edunov, and Scialom}]{touvron2023llama}
Hugo Touvron, Louis Martin, Kevin Stone, Peter Albert, Amjad Almahairi, Yasmine Babaei, Nikolay Bashlykov, Soumya Batra, Prajjwal Bhargava, Shruti Bhosale, Dan Bikel, Lukas Blecher, Cristian~Canton Ferrer, Moya Chen, Guillem Cucurull, David Esiobu, Jude Fernandes, Jeremy Fu, Wenyin Fu, Brian Fuller, Cynthia Gao, Vedanuj Goswami, Naman Goyal, Anthony Hartshorn, Saghar Hosseini, Rui Hou, Hakan Inan, Marcin Kardas, Viktor Kerkez, Madian Khabsa, Isabel Kloumann, Artem Korenev, Punit~Singh Koura, Marie-Anne Lachaux, Thibaut Lavril, Jenya Lee, Diana Liskovich, Yinghai Lu, Yuning Mao, Xavier Martinet, Todor Mihaylov, Pushkar Mishra, Igor Molybog, Yixin Nie, Andrew Poulton, Jeremy Reizenstein, Rashi Rungta, Kalyan Saladi, Alan Schelten, Ruan Silva, Eric~Michael Smith, Ranjan Subramanian, Xiaoqing~Ellen Tan, Binh Tang, Ross Taylor, Adina Williams, Jian~Xiang Kuan, Puxin Xu, Zheng Yan, Iliyan Zarov, Yuchen Zhang, Angela Fan, Melanie Kambadur, Sharan Narang, Aurelien Rodriguez, Robert Stojnic, Sergey Edunov, and Thomas
  Scialom. 2023.
\newblock \href {http://arxiv.org/abs/2307.09288} {Llama 2: Open foundation and fine-tuned chat models}.

\bibitem[{Wang et~al.(2023)Wang, Zhang, Wang, Lim, and Wang}]{wang2023llm4visexplainablevisualizationrecommendation}
Lei Wang, Songheng Zhang, Yun Wang, Ee-Peng Lim, and Yong Wang. 2023.
\newblock \href {http://arxiv.org/abs/2310.07652} {Llm4vis: Explainable visualization recommendation using chatgpt}.

\bibitem[{Xia et~al.(2025)Xia, Zhang, Ye, Yan, Liu, Zhou, Chen, Dou, Shi, Yan, and Qiao}]{xia2025chartxchartvlmversatile}
Renqiu Xia, Bo~Zhang, Hancheng Ye, Xiangchao Yan, Qi~Liu, Hongbin Zhou, Zijun Chen, Min Dou, Botian Shi, Junchi Yan, and Yu~Qiao. 2025.
\newblock \href {http://arxiv.org/abs/2402.12185} {Chartx \& chartvlm: A versatile benchmark and foundation model for complicated chart reasoning}.

\bibitem[{Zadeh et~al.(2024)Zadeh, Kim, Kim, and Kim}]{zadeh2024text2chart31instructiontuningchart}
Fatemeh~Pesaran Zadeh, Juyeon Kim, Jin-Hwa Kim, and Gunhee Kim. 2024.
\newblock \href {http://arxiv.org/abs/2410.04064} {Text2chart31: Instruction tuning for chart generation with automatic feedback}.

\bibitem[{Zhang et~al.(2024)Zhang, Wang, Li, Shen, Cao, and Wang}]{zhang2024chartifytextautomatedchartgeneration}
Songheng Zhang, Lei Wang, Toby Jia-Jun Li, Qiaomu Shen, Yixin Cao, and Yong Wang. 2024.
\newblock \href {http://arxiv.org/abs/2410.14331} {Chartifytext: Automated chart generation from data-involved texts via llm}.

\bibitem[{Zhu et~al.(2024)Zhu, Yuan, Wang, Liu, Liu, Deng, Chen, Liu, Dou, and Wen}]{zhu2024largelanguagemodelsinformation}
Yutao Zhu, Huaying Yuan, Shuting Wang, Jiongnan Liu, Wenhan Liu, Chenlong Deng, Haonan Chen, Zheng Liu, Zhicheng Dou, and Ji-Rong Wen. 2024.
\newblock \href {http://arxiv.org/abs/2308.07107} {Large language models for information retrieval: A survey}.

\end{thebibliography}

\appendix
\newpage
\section{Appendix}
\label{sec:appendix}
\subsection{Qualitative examples}
Intent: Examine the average realized prices per barrel of oil in the United States, International markets, and globally for the years 2021-2023.
\begin{figure}[ht]
    \centering    \includegraphics[width=0.5\textwidth]{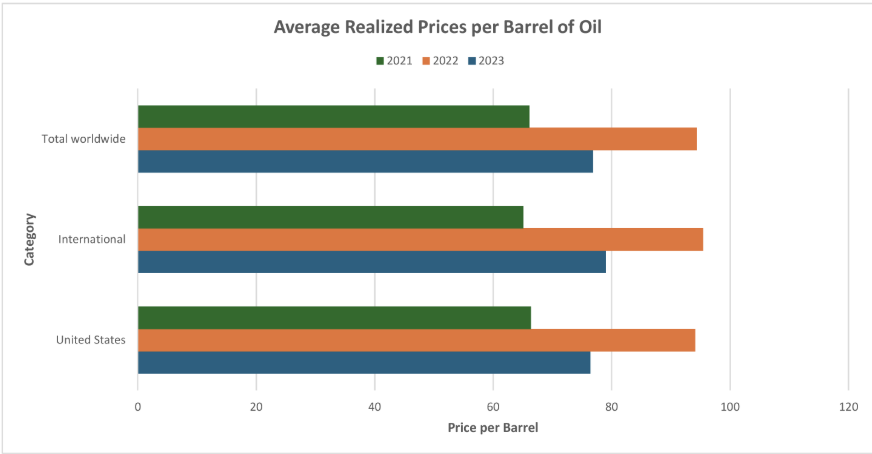}
    \caption{Ground Truth Chart}
    \label{fig:sample_GT_chart}
\end{figure}
\begin{figure}[ht]
    \centering    \includegraphics[width=0.5\textwidth]{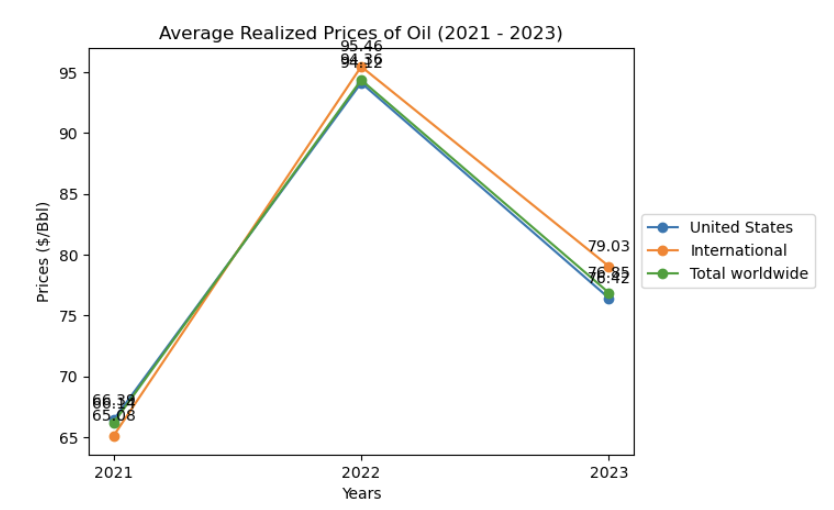}
    \caption{Baseline - Generated Chart}
    \label{fig:sample_baseline_chart}
\end{figure}
\begin{figure}[ht]
    \centering    \includegraphics[width=0.5\textwidth]{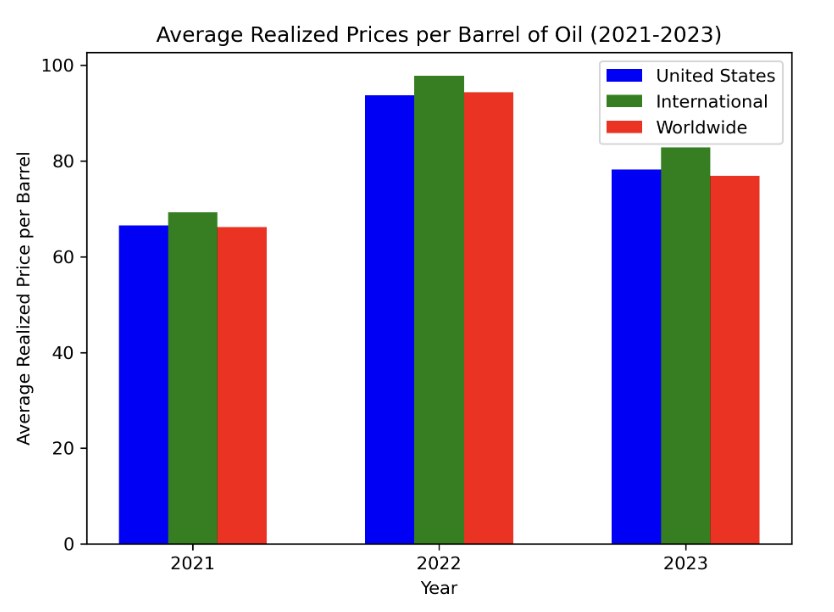}
    \caption{Chart generated by our proposed approach}
    \label{fig:sample_our_chart}
\end{figure}

\clearpage
\begin{figure*}[t]
\centering
\begin{mdframed}[backgroundcolor=gray!10, roundcorner=5pt]
\small
\textbf{Task:} \\
Extract structured chart data from the provided content based on the user's intent, adhering to the specified JSON format.

\textbf{Input:}
\begin{itemize}
    \item User Intent: \texttt{\{intent\}}
    \item Content: \texttt{\{content\}}
    \item Optional Feedback (if available): \texttt{\{optional\_feedback\_section\}}
    \item Output Format Schema: \texttt{\{output\_format\}}
\end{itemize}

\textbf{Instructions:}
\begin{enumerate}
    \item Carefully read the User Intent.
    \item \textbf{Internal Thought Process (Mentally follow these steps):}
    \begin{itemize}
        \item \textbf{Decompose:} Break down the intent into specific data points, labels, categories, and title.
        \item \textbf{Locate:} Scan the content for exact data matching the above.
        \item \textbf{Extract \& Structure:} Collect and format data strictly according to the schema.
    \end{itemize}
    \item Extract relevant data points: $(x, y, \text{category})$, axis labels, and chart title.
    \item \textbf{If feedback is provided:} Focus on fixing issues like missing elements or ignored sections. Adjust your decomposition and extraction accordingly.
    \item Output must follow the JSON schema exactly. Keep numeric formats consistent.
    \item Output only the JSON object. \textbf{Do not} include explanations or markdown like \texttt{```json}.
\end{enumerate}

\textbf{Example Output Format:}
\begin{verbatim}
{
  "values": [
    {
      "x": "[string or number]",
      "y": "[number or string representing number]",
      "category": "[string, optional]"
    }
  ],
  "x_axis_label": "[string]",
  "y_axis_label": "[string]",
  "title": "[string]"
}
\end{verbatim}
\end{mdframed}
\caption{Chart Data Extraction}
\label{fig:intent-breakdown}
\end{figure*}

\begin{figure*}[t]
\centering
\begin{mdframed}[backgroundcolor=gray!10, roundcorner=5pt]
\small
\textbf{Task:} Validate the extracted chart data against the source content and user intent. Determine if re-extraction is necessary or if only minor refinements are needed.

\textbf{Input:}
\begin{itemize}
    \item Original Intent: \texttt{\{intent\}}
    \item Source Content: \texttt{\{content\}}
    \item Extracted Chart Data: \texttt{\{extracted\_data\}} \textit{// JSON object from the extraction step}
    \item Expected Schema: \texttt{\{output\_format\}}
\end{itemize}

\textbf{Validation Checks to Perform:}
\begin{enumerate}
    \item \textbf{Intent Fulfillment \& Source Coverage:} Does the \texttt{extracted\_data} capture the key information requested in the \texttt{intent} that is present in the \texttt{Source Content}? Are there critical omissions?
    \item \textbf{Data Accuracy:} Are the values (\texttt{x}, \texttt{y}, \texttt{category}) and labels/title in \texttt{extracted\_data} accurately reflecting the \texttt{Source Content}?
\end{enumerate}

\textbf{Response Format:}
\begin{verbatim}
{
  "needs_re_extraction": "[true/false]",
  "feedback_for_re_extraction": "[string]",
  "suggested_corrections_for_refinement": [
    {
      "field_path": "[JSON path, e.g., values[0].y or title]",
      "suggestion": "[Brief description of the fix, e.g., 'Convert to number']",
      "suggested_value": "[Optional: The corrected value if easily determined]"
    }
  ],
  "confidence_score": "[0-10 score reflecting confidence in the data]"
}
\end{verbatim}

Focus on the primary decision: re-extract or refine/accept. Keep feedback concise. \\
Output only a valid JSON and no other text. Do not add prefix like \texttt{```json...}
\end{mdframed}
\caption{Chart Data Validation}
\label{fig:validate-chart-data}
\end{figure*}

\begin{figure*}[t]
\centering
\begin{mdframed}[backgroundcolor=gray!10, roundcorner=5pt]
\small
\textbf{Task:} Apply the suggested minor corrections to the extracted chart data.

\textbf{Input:}
\begin{itemize}
    \item Original Intent: \texttt{\{intent\}} 
    \item Source Content: \texttt{\{content\}} 
    \item Extracted Data (Pre-Refinement): \texttt{\{extracted\_data\}} 
    \item Suggested Corrections: \texttt{\{suggested\_corrections\}} 
    \item Expected Schema: \texttt{\{output\_format\}}
\end{itemize}

\textbf{Instructions:}
\begin{enumerate}
    \item Iterate through the \texttt{Suggested Corrections}.
    \item Apply each correction to the corresponding \texttt{field\_path} in the \texttt{Extracted Data}. Use \texttt{suggested\_value} if provided, otherwise interpret the \texttt{suggestion}.
    \item Ensure the final \texttt{refined\_data} strictly follows the \texttt{Expected Schema} provided in the input.
    \item Do \textbf{not} add new data or make changes beyond the \texttt{Suggested Corrections}.
\end{enumerate}

\textbf{Response Format:}
\begin{verbatim}
{
  "refined_data": { /* The data structure with corrections applied, adhering to the Expected Schema */ },
  "refinement_summary": {
    "changes_applied_count": "[number]", // Count of corrections successfully applied
    "issues_applying_corrections": ["[List any suggestions that could not be applied and why]"]
  }
}
\end{verbatim}

Output only a valid JSON and no other text. Do not add prefix like ```json...```
\end{mdframed}
\caption{Chart Data Refinement}
\label{fig:refine-chart-data}
\end{figure*}

\begin{figure*}[t]
\centering
\begin{mdframed}[backgroundcolor=gray!10, roundcorner=5pt]
\small
\textbf{Task:} \\
Evaluate and recommend statistical chart visualizations based on the intent and the final, validated (and potentially refined) data.

\textbf{Input:}
\begin{itemize}
    \item Intent: \texttt{\{intent\}}
    \item Final Chart Data: \texttt{\{data\}}
\end{itemize}

\textbf{Heuristics Framework:} \\
Consider these guidelines:
\begin{itemize}
    \item \textbf{Time-based:} $\leq$3 points $\rightarrow$ Bar; 4+ points $\rightarrow$ Line; Irregular spacing $\rightarrow$ Grouped Bar
\item \textbf{Comparison:} Few categories (2--5) $\rightarrow$ Bar; Many (6+) $\rightarrow$ Stacked Bar; Proportions $\rightarrow$ Pie ($\leq$6 segments)

    \item \textbf{Intent:} Magnitude $\rightarrow$ Bar; Trend $\rightarrow$ Line; Composition $\rightarrow$ Pie/Stacked Bar
    \item \textbf{Anti-Patterns:} Avoid cluttered pies and sparse lines. Prioritize readability.
\end{itemize}

\textbf{Requirements:}
\begin{enumerate}
    \item Analyze the structure and nature of the \texttt{Final Chart Data} (types of x/y values, number of points/categories).
    \item Relate the data structure to the \texttt{Intent}.
    \item Evaluate potential chart types based on the heuristics.
    \item Recommend the most appropriate chart type with justification.
\end{enumerate}

\textbf{Response Format:}
\begin{verbatim}
{
    "recommended_chart_type": "[Best-suited chart type]",
    "justification": "[Reason based on data structure, intent, and heuristics]",
    "confidence_score": "[0-10 score indicating recommendation strength]"
}
\end{verbatim}
Output only a valid JSON and no other text. Do not add a prefix like ```json.
\end{mdframed}

\caption{Chart Type Prediction}
\label{fig:chart-type-prediction}
\end{figure*}
\end{document}